\documentclass[a4paper, 10pt, conference]{ieeeconf}      % Use this line for a4
\usepackage{FG2021}

\FGfinalcopy % *** Uncomment this line for the final submission

\IEEEoverridecommandlockouts                              % This command is only
\usepackage{graphics} % for pdf, bitmapped graphics files
\usepackage{epsfig} % for postscript graphics files
\usepackage{mathptmx} % assumes new font selection scheme installed
\usepackage{times} % assumes new font selection scheme installed
\usepackage{amsmath} % assumes amsmath package installed
\usepackage{amssymb}  % assumes amsmath package installed
\usepackage{multirow}
\usepackage{subcaption}

\newcommand{\etal}{\textit{et al.}}

\def\FGPaperID{****} % *** Enter the FG2021 Paper ID here

\title{\LARGE \bf
Partial Attack Supervision and Regional Weighted Inference for Masked Face Presentation Attack Detection
}

\author{\parbox{16cm}{\centering
    {\large Meiling Fang$^{1,2}$, Fadi Boutros$^{1,2}$, Arjan Kuijper$^{1,2}$, Naser Damer$^{1,2}$}\\
    {\normalsize
    $^{1}$Fraunhofer Institute for Computer Graphics Research IGD,
        Darmstadt, Germany\\
    $^{2}$Department of Computer Science, TU Darmstadt, Darmstadt, Germany}}
    \thanks{This research work has been funded by the German Federal Ministry of Education and Research and the Hessian Ministry of Higher Education, Research, Science and the Arts within their joint support of the National Research Center for Applied Cybersecurity ATHENE.}% <-this % stops a space
}

\begin{document}
%%%%%%%%%%%%%%
% COPYRIGHT NOTICE - Uncomment correct version below
%
% The notices are from the FG 2021 LOA 
%
% Active is the "Others" option - see Case #4 in the instructions posted at: http://iab-rubric.org/fg2021
%
%%%%%%%%%%%%%%

% Case #1: For papers in which all authors are employed by the US government, the copyright notice is: 
%\IEEEoverridecommandlockouts\pubid{\makebox[\columnwidth]{U.S. Government work not protected by U.S. copyright \hfill}
%\hspace{\columnsep}\makebox[\columnwidth]{ }}

% Case #2: For papers in which all authors are employed by a Crown government (UK, Canada, and Australia), the copyright notice is:
%\IEEEoverridecommandlockouts\pubid{\makebox[\columnwidth]{978-1-6654-3176-7/21/\$31.00~\copyright{}2021 Crown \hfill}
%\hspace{\columnsep}\makebox[\columnwidth]{ }}

% Case #3: For papers in which all authors are employed by the European Union, the copyright notice is:
%\IEEEoverridecommandlockouts\pubid{\makebox[\columnwidth]{978-1-6654-3176-7/21/\$31.00~\copyright{}2021 European Union \hfill}
%\hspace{\columnsep}\makebox[\columnwidth]{ }}

% Case #4: For all other papers the copyright notice is:
\IEEEoverridecommandlockouts\pubid{\makebox[\columnwidth]{978-1-6654-3176-7/21/\$31.00~\copyright{}2021 IEEE \hfill}
\hspace{\columnsep}\makebox[\columnwidth]{ }}

\maketitle

%%%%%%%%%%%%%%%%%%%%%%%%%%%%%%%%%%%%%%%%%%%%%%%%%%%%%%%%%%%%%%%%%%%%%%%%%%%%%%%%
\begin{abstract} % ready for Naser
Wearing a mask has proven to be one of the most effective ways to prevent the transmission of SARS-CoV-2 coronavirus. However, wearing a mask poses challenges for different face recognition tasks and raises concerns about the performance of masked face presentation detection (PAD).
The main issues facing the mask face PAD are the wrongly classified bona fide masked faces and the wrongly classified partial attacks (covered by real masks). This work addresses these issues by proposing a method that considers partial attack labels to supervise the PAD model training, as well as regional weighted inference to further improve the PAD performance by varying the focus on different facial areas. 
Our proposed method is not directly linked to specific network architecture and thus can be directly incorporated into any common or custom-designed network. In our work, two neural networks (DeepPixBis \cite{deeppix_19} and MixFaceNet \cite{mixfacenet}) are selected as backbones. 
The experiments are demonstrated on the collaborative real mask attack (CRMA) database \cite{masked_face_pad}. Our proposed method outperforms established PAD methods in the CRMA database by reducing the mentioned shortcomings when facing masked faces. 
Moreover, we present a detailed step-wise ablation study pointing out the individual and joint benefits of the proposed concepts on the overall PAD performance.
\end{abstract}

%%%%%%%%%%%%%%%%%%%%%%%%%%%%%%%%%%%%%%%%%%%%%%%%%%%%%%%%%%%%%%%%%%%%%%%%%%%%%%%%
\section{INTRODUCTION} % ready for Naser
\begin{figure*}[htb]
\begin{center}
\includegraphics[width=0.95\linewidth]{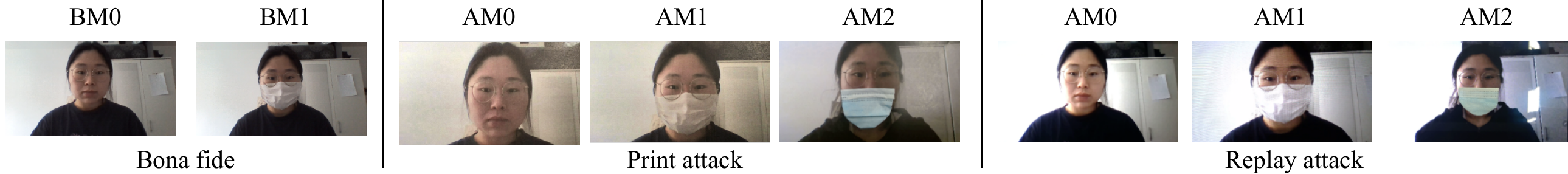}
\end{center}
\caption{Samples from the CRMA database \cite{masked_face_pad}. The first two images are bona fide samples with subjects without masks (BM0) and with face masks (BM1). Based on such bona fide samples, print and replay attacks were created. In addition to entirely unmasked (AM0) and masked (AM1) print or replay attacks, a partial attack (AM2) was produced, where a real face mask was placed on unmasked spoof faces.}
\label{fig:image_examples}
\end{figure*}

In recent years, face biometrics has become widely used and applied in many scenarios, such as unlocking mobile phones and automated border control (ABC). As a result, face recognition (FR) research \cite{DBLP:conf/icpr/StrucDP10,mixfacenet,elasticface} has shown significant progress over the past decade. Meanwhile, face presentation attack detection (PAD) has also attracted more and more attention due to the wide application of FR systems. PAD aims at securing the FR systems from presentation attacks (PAs), such as printed photos and replayed videos. Attackers can use such PAs to spoof FR systems by impersonating someone or obfuscating their identity. Many works leverage deep learning techniques and made a remarkable improvement in FR and PAD problems.    
However, the recent COVID-19 pandemic rendered the conventional FR and PAD solutions less effective in many cases as face masks present FR/PAD algorithms with unexpected face presentation. Damer et al. \cite{DBLP:conf/biosig/DamerGCBKK20,IET_B_Mask} studied the effect of face mask on the performance of FR verification. Their experimental results have shown that FR algorithms designed before the COVID-19 pandemic suffer performance degradation owing to the masked faces. A follow up study showed that this effect extends even to verification decisions made by human operators \cite{DBLP:journals/corr/abs-2103-01924}. Subsequently, many methods have been developed to target the masked FR problem. For example, several works proposed to train FR models by adding masked face data or simulated masked faces \cite{DBLP:conf/icb/BoutrosDKRKRKFZ21,DBLP:journals/Anwar,Ekrem_biosig21} or train models to focus on the unmasked regions \cite{DBLP:journals/apin/LiGLL21,PedroBiosig}. Moreover, Boutros et al. \cite{fadi_pr21} proposed embedding unmasking model (EUM) operated on the top of existing face recognition models and trained using the self-restrained triplet loss function to enable the EUM to produce embeddings similar to these of unmasked faces. Their proposed method reduced the negative impact of wearing face masks on FR performance. Despite much attention paid to the masked FR problem, masked PAD is still understudied. Fang \cite{masked_face_pad} et al. presented a collaborative real mask attack (CRMA) database containing three types of PAs, the unmasked print/replay attack (AM0), masked print/replay attack (AM1), and partially masked attack where spoof faces are partially covered by real masks (AM2). Figure \ref{fig:image_examples} shows samples of the CRMA database. They conducted extensive experiments to explore the effect of masked bona fide, masked attacks, and partially masked attacks on the face PAD behavior. Their experimental results indicated that masked bona fide and PAs dramatically decreased the performance of PAD algorithms. Furthermore, they showed that deep-learning-based methods performed worse on the partially masked attack (AM2) than the masked attack (AM1) in most cases.
Nevertheless, in their work \cite{masked_face_pad}, only the effect of masked PAs on the PAD and FR performance was investigated by utilizing several PAD algorithms designed before the COVID-19 pandemic, no solutions were proposed to target the challenges raised by the masked faces. 
Therefore, to address the issue of masked face PAD, especially partially masked attacks, we introduce a solution that combines two novel modules, partial attack label (PAL) and regional weighted inference (RW). The PAL module is inspired by the pixel-wise supervision \cite{deeppix_19, schuffled_pixbis, Liu_auxiliary_siw_18}. However, unlike using a coherent map as the ground truth of partial attack (AM2) in \cite{deeppix_19}, we propose annotating the partially covered real mask region as bona fide. The fine-grained partial attack label aims to enable better supervision during model training. Once the model is trained, the RW is used in the inference phase for further PAD decision optimization. The regional weighted inference is inspired by previous observations in \cite{masked_face_pad, fu_biosig_2021, fu_biosig_morph_2021} stating that the eye region contributes more significantly in different face-related tasks, such as PAD \cite{masked_face_pad} or face image quality assessment \cite{fu_biosig_2021,DBLP:journals/corr/abs-2110-11283}. Based on this assumption, we weigh different regions of the predicted feature map and thus enhance the performance of the PAD decision. 

The main contributions of this work are: 1) a PAD method is proposed to target the masked face PAD limitations based on two novel components, the partial attack label supervision and the regional weighted inference (PAL-RW); 2) a detailed step-wise ablation experiments is conducted, showing the benefits induced by our PAL and RW modules, respectively; 3) the results demonstrated on the CRMA database indicate an improved performance in the perspective of a set of established PAD solutions. In addition, conceptually,  our proposed PAL-RW method focuses on fine-graining training ground truth information and post-processing predictions. Thus, it can be simply incorporated into any common or specially designed neural network architecture.

We introduce the relevant works in Section \ref{sec:related_work}. Then, our PAL-RW method is described in detail in Section \ref{sec:method}. Section \ref{sec:results} introduces the used database, experimental settings, evaluation metrics, and then discusses the PAD results. Finally, conclusions are presented in Section \ref{sec:conclusion}.

%%%%%%%%%%%%%%%%%%%%%%%%%%%%%%%%%%%%%%%%%%%%%%%%%%%%%%%%%%%%%%%%%%%%%%%%%%%%%%%%
\section{Related work} % ready for Naser
\label{sec:related_work}
In this section, we review existing relevant works including CNN-based algorithms with training supervision strategies along with recent face PAD databases.  

\subsection{Presentation attack detection}
% current pixel-wise supervision methods, goals, pros and cons
Early face PAD methods utilized motion information \cite{DBLP:conf/bmvc/DamerD16} (e.g., eye blinking, head motion), handcrafted features (e.g., Local Binary Patterns (LBP) \cite{DBLP:journals/pami/OjalaPM02}, or 2D Fourier Spectrum \cite{DBLP:conf/eccv/JourablooLL18}) to detect presentation attack. Recently, Deep learning-based methods have shown significant improvement in many general computer vision fields, such as classification or object detection. As a result, many face PAD works employ Convolutional Neural Network (CNN) techniques. These CNN-based PAD works can be grouped into two classes based on the used training strategy: 1) global binary supervision where a network is supervised by a binary scalar label while training \cite{FASNet,pad_competition}, and 2) pixel-wise supervision where \cite{deeppix_19,schuffled_pixbis,DBLP:journals/tifs/DebJ21,DBLP:journals/tbbis/YuLSXZ21,FangWACV2022} a network outputs a feature map and the training is supervised (at least partially) by pixel-wise mask. George and Marcel \cite{deeppix_19} first introduced the deep pixel-wise binary supervision (DeepPixBis) method, which outputs an intermediate feature map with the size of $14 \times 14$ pixels to assist the binary classification. 
This feature map was considered as the prediction scores generated from the patches in the input image. 
The $14 \times 14$ pixel-wise label used for training supervision is set to zero for attack and to one for bona fide samples. However, the performance degradation might be caused by the noisy pixel-wise label when handling partial attacks. Kantarci \etal \cite{schuffled_pixbis} improved the performance of the DeepPixBis method by shuffling the patches from input images and combining face patches. The generated new inputs improved the generalizability of the trained model under cross-database scenarios. Yu \etal \cite{DBLP:journals/tbbis/YuLSXZ21} proposed a pyramid pixel-wise supervision method, which decomposed the pixel-wise label into multiple spatial scales for the supervision of multi-scale deep features. The pyramid supervision is able to interpret a richer spatial context, which is beneficial for fine-grained feature learning. 
Overall, pixel-wise labels have proven to be helpful in the improvement of PAD performance. Nevertheless, the quality of the pixel-wise labels is essential for the convergence of the trained networks. The coarse all zero or all one pixel-wise binary mask might not be suitable for the partial attack, which is the issue addressed in this work. 

%In supervised/meta-supervised deep learning, data label is the key to guide the model to learn informative features. In PAD field, the label of an image is either bona fide or attack.
% the pixel-wise supervision need strict and exhausted pixel-wise annotated labels.

\subsection{Masked faces and presentation attacks}
% current datasets, focus on partial attacks and Collaborative Real Mask Attack databases (CRMA), goal and problems
In addition to the quality of the labels, data resources are also crucial for model generalization. The insufficient (amount or variation) data might cause the underfitting or overfitting of the PAD network. Moreover, the lack of diversity in the data impacts the generalization ability of the model. Several face PAD databases were previously released and contributed to the significant progress of PAD research. For example, CASIA-FAS \cite{casia_fas}, MSU-MFSD \cite{msu_mfs}, OULU-NPU \cite{oulu_npu} databases comprise of 2D print and replay attack.
Furthermore, to fit the ongoing COVID-19 pandemic, the CRMA database \cite{masked_face_pad} is introduced, containing print and replay attacks where subjects were wearing face masks (AM1) and partial attack that a real mask was placed on printed photos or screens running replayed videos (more details can be found in Section \ref{ssec:database}). Figure \ref{fig:image_examples} presents the different bona fide and attack samples in the CRAM database. AM2 is the partial attack where the face mask is real, and the uncovered face region is an attack. In \cite{masked_face_pad}, the effect of face mask on PAD performance was explored by comparing the results of several PAD methods, including DeepPixBis \cite{deeppix_19}. This study has shown that the performance of different PAD solutions is strongly affected when processing masked faces. However, solutions utilizing pixel-wise ground truth of the attack sample in their work was a zero map that did not consider partial attack labels. The noisy ground truth might affect the training process of the model and thus cause performance degradation. 

To overcome the issues on partially masked attacks and masked face PAD in general, our proposed solution considers partial pixel-wise labeling and varying consideration of different facial areas.

%%%%%%%%%%%%%%%%%%%%%%%%%%%%%%%%%%%%%%%%%%%%%%%%%%%%%%%%%%%%%%%%%%%%%%%%%%%%%%%%
\section{Methodology} % NAser DONE
\label{sec:method}

\begin{figure*}[htb]
\begin{center}
\includegraphics[width=0.73\linewidth]{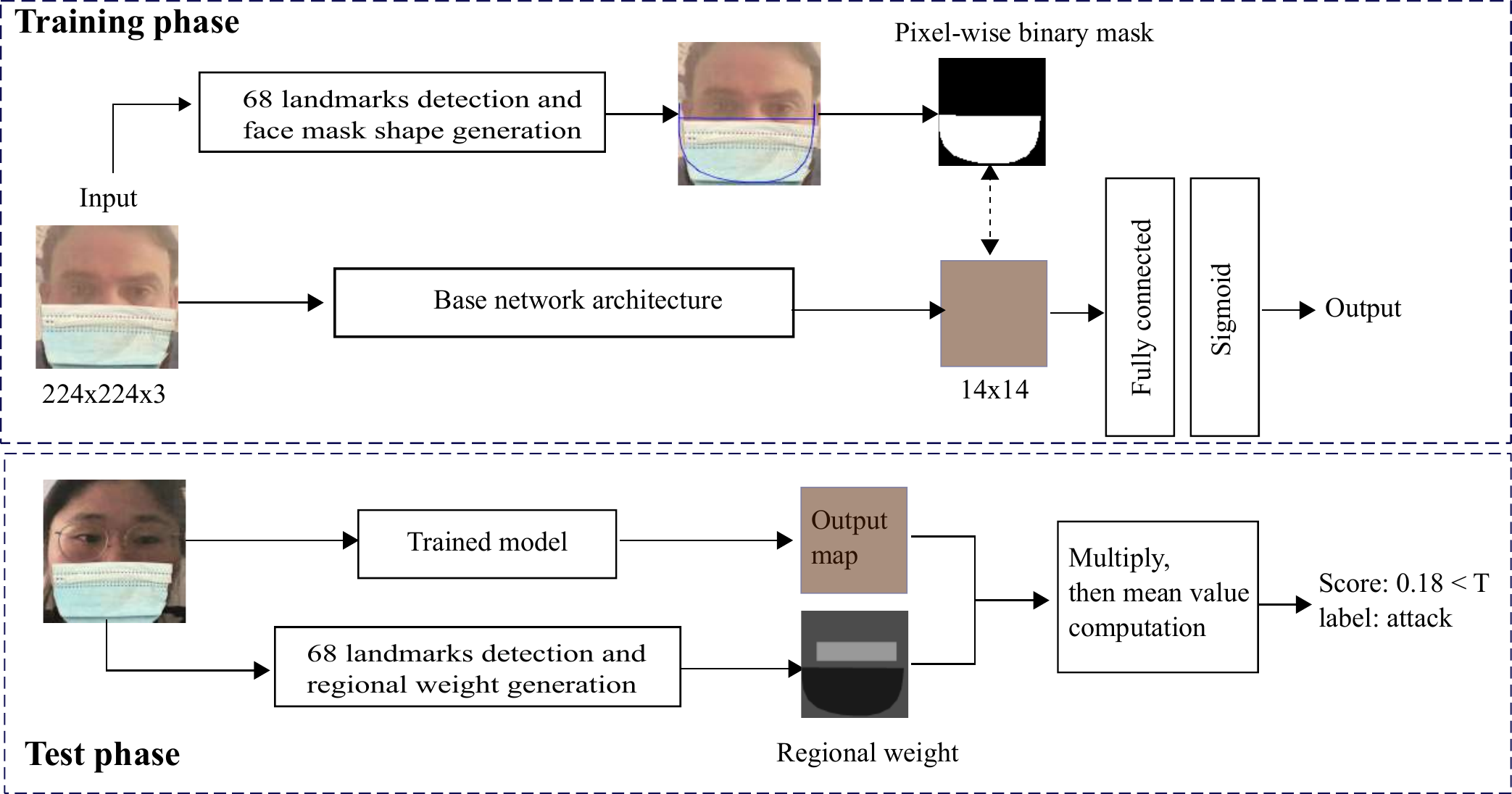}
\end{center}
\caption{An overview of our proposed PAL-RW method. The input of the model is a face image of $224 \times 224 \times 3$ pixels, and the model is supervised by a feature map with the size of $14 \times 14$ in addition to a binary output. In the training phase, the pixel-wise label for partial attack (AM2) is produced based on 68 facial landmarks. The pixel-wise label of the real face mask in AM2 data is set to 1 (bona fide) instead of a zero map for this attack. In the inference phase, the final PAD decision score is the mean value of the regional weighted feature map for further performance improvement. The lighter the color in the region weight map, the higher the weight value, that is, the eye region contributes more to PAD decision.}
\label{fig:method}
\vspace{-5mm}
\end{figure*}

% add we focus on the CRMA
This section describes the proposed method (PAL-RW) that utilizes our partial attack label supervision (PAL) and regional weighted inference (RW) for masked face PAD, along with the used backbone network architectures. The PAL intends to provide more accurate ground truth for partial attacks and thus enhances the convergence of the model training. Moreover, the RW post-processes the prediction results by increasing the focus on certain facial areas to further improve the PAD performance. Figure \ref{fig:method} depicts the training and testing phase in detail.

% 1. ground truth improvement, image with face mask/ 

\subsection{Partial Attack Label}
\label{ssec:pal}
\begin{figure*}[htb]
\centering
\subfloat[Binary scalar label]{
	\label{subfig:labels_1}
	\includegraphics[width=0.25\textwidth]{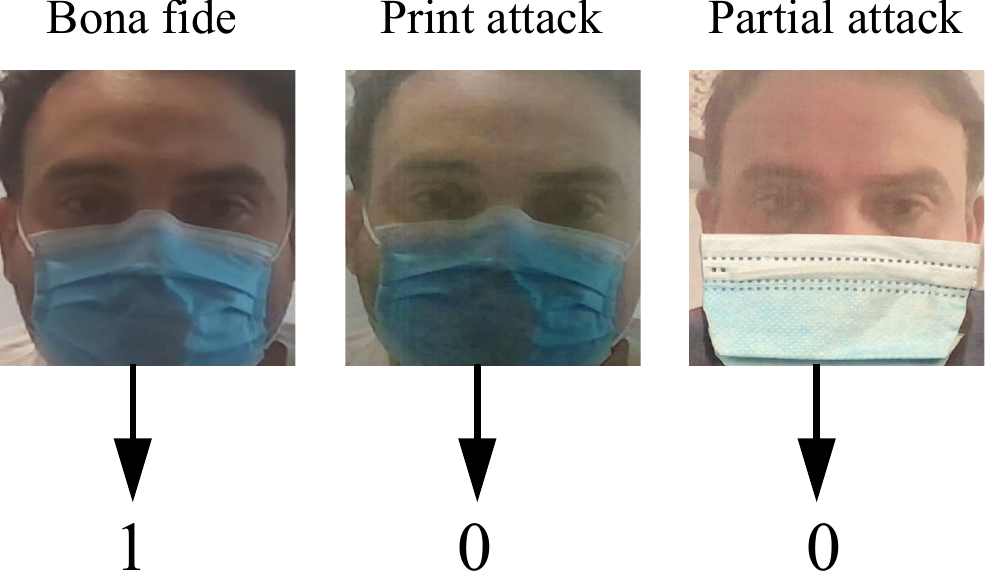}} 
\subfloat[Binary Mask used in \cite{masked_face_pad}]{
	\label{subfig:labels_2}
	\includegraphics[width=0.25\textwidth]{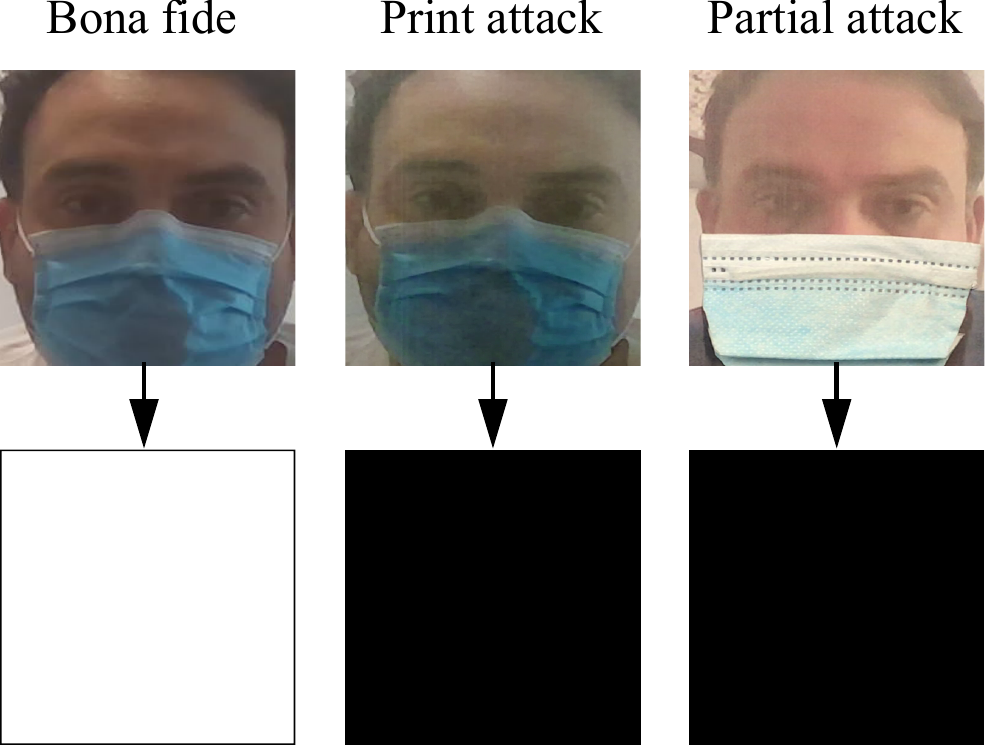}} 
\subfloat[Our partial attack label]{
	\label{subfig:labels_3}
	\includegraphics[width=0.25\textwidth]{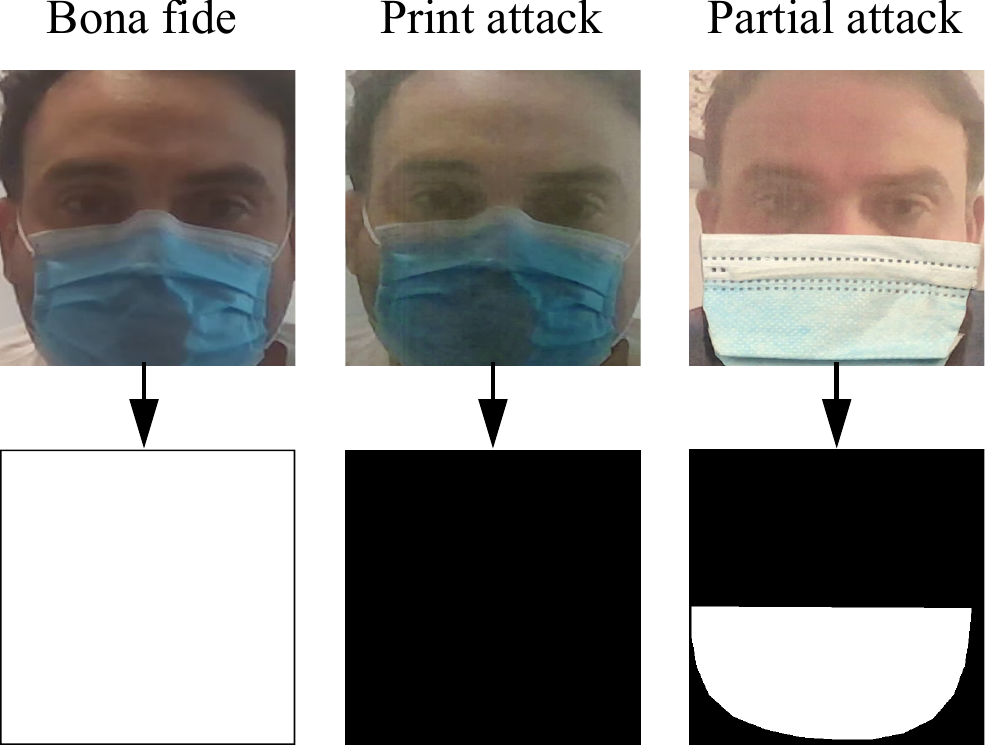}}
\caption{Examples of three supervision methods. (a) Binary scalar labels for global supervision. For example, 1 represents bona fide and 0 represents attack \cite{FASNet, pad_lbp_2011}. (b) Pixel-wise binary mask for local supervision such as used in \cite{deeppix_19, masked_face_pad}. A zero map is usually treated as ground truth of attack and a one map is for bona fide. (c) Our proposed partial attack label for AM2 attacks. The real face mask placed on an attack face is segmented and annotated as bona fide, while the rest of the attack is labeld as an attack.  }
\label{fig:labels}
\vspace{-5mm}
\end{figure*}

Recent PAD works can be grouped into two classes: 1) global binary supervision where a network is supervised by a binary scalar label while training \cite{FASNet,pad_competition} (as shown in Figure \ref{subfig:labels_1}), and 2) pixel-wise binary supervision where a network outputs a feature map and is supervised by binary mask, such as \cite{deeppix_19,schuffled_pixbis,DBLP:journals/tifs/DebJ21,DBLP:journals/tbbis/YuLSXZ21}. Each pixel in the feature map is assumed as either bona fide (1) or attack (0). Most pixel-wise supervision methods \cite{deeppix_19,schuffled_pixbis} utilized a zero map as ground truth of attacks and a one map for bona fide samples as shown in Figure \ref{subfig:labels_2}). They improved the performance on print and replay attacks, but performance degradation might occur when dealing with partial attacks. A possible reason is the incorrect pixel-wise labels of attacks. For example, the CRMA database \cite{masked_face_pad} contains an attack (AM2 in Figure \ref{fig:image_examples}) where a real mask was placed on unmasked spoof faces. For exploring the effect of face masks on PAD performance, Fang \etal adopted DeepPixBis \cite{deeppix_19} method. However, in their work, the binary mask for this partially masked face attack was set completely to zero. To improve the performance, we intend to make the pixel-wise label more accurate and thus enhance the training by partial attack label supervision. As shown in Figure \ref{subfig:labels_3}, the shape of the face mask is detected based on 68 facial landmarks, and each pixel-wise label in the face mask region is set to one (bona fide). Note that this operation on the binary mask is only performed for the partially masked face attack (AM2).

% 2. regional weight based on observation of score-cam, weight image
\subsection{Model Architecture}
\label{ssec:model}
To train a PAD solution based on our partial attack labels, we select two network architectures (DeepPixBis \cite{deeppix_19} and MixFaceNet \cite{mixfacenet}) as our backbones to validate our proposed PAL and RW on multiple architectures.
DeepPixBis \cite{deeppix_19} model was used for exploring the effect of face masks in the CRMA database \cite{masked_face_pad} and achieved good performance, relative to other baselines. Therefore, we choose DeepPixBis as one of our backbone models for further comparison with results reported in \cite{masked_face_pad}.
MixFaceNet \cite{mixfacenet} is chosen as the second utilized backbone architecture because it is an extremely efficient architecture for face verification and identification, which possesses lower computation complexity (FLOPs) and high accuracy. MixFaceNet additionally contains different sizes of convolutional kernels, which might be beneficial to capture different levels of attack clues. 
The MixFaceNet is used with the weights initialized as the publically available weights pre-trained on the MS1MV2 dataset for face recognition, which might help maintain more subtle facial features.
On the other hand, the DeepPixBis \cite{deeppix_19} architecture is used with the initial weights set by pre-training on the ImageNet dataset \cite{ImageNet} for the general computer vision tasks, as described in \cite{deeppix_19}. 
DeepPixBis \cite{deeppix_19} method uses DenseNet \cite{densenet} as a base network architecture and outputs a feature map of $14 \times 14$ pixels and a binary scalar prediction. The training of DeepPixBis is supervised by a binary mask and a binary label. 
MixFaceNet was partially inspired by the MixNets, and the channel shuffle operation to the MixConv block was introduced for enhancing the FR performance. In our case, we change the input image size of $112 \times 112 \times 3$ to $224 \times 224 \times 3$ for outputting a feature map of $14 \times 14$ pixels, to be identical to the DeepPixBis backbone. Furthermore, the embedding stage of MixFaceNet is removed and replaced by two fully connected layers. The first fully connected layer is employed to output a feature map for pixel-wise supervision, while the second fully connected layer is used for binary classification. The MixFaceNet is also supervised by a binary mask and a binary label.
Both models are trained by Binary Cross Entropy (BCE) loss function. 
\begin{equation}
    \mathcal{L}_{BCE} = -[y \cdot \log p + (1-y)\cdot \log (1-p)] ,
\end{equation}
where y is the ground truth (1 for bona fide and 0 for attack in our case) and p is predicted probability. 
The overall loss equation for training of both models is shown below:
\begin{equation}
    \mathcal{L}_{overall} = \lambda \cdot \mathcal{L}_{BCE}^{pixel-wise} + (1 - \lambda) \cdot \mathcal{L}_{BCE}^{binary} ,
\end{equation}
where $\lambda$ is set to 0.5 in our experiments.

\subsection{Regional Weighted Inference}
\label{ssec:rw}
% post-processing
Once the model is trained, we can estimate the probability of a given image is an attack or bona fide. Most pixel-wise supervision-based methods use the mean value of the output feature map as the final decision score \cite{deeppix_19,DBLP:conf/icb/FangDBKK21}, which neglects the differences in features at various facial regions. Therefore, we propose the regional weighted inference to post-process the prediction scores of models. The RW is motivated by two main previous observations: 1) Fang \etal \cite{masked_face_pad} found that the eye regions (including eyebrows) had a more significant influence on the PAD prediction probability than the face mask and other facial regions. 2) Fu \etal\cite{fu_biosig_2021} explored the contributions of different face sub-regions to the face image quality and their experiments indicated that the eye region quality largely affects FR performance and additionally shows consistent quality degradation in face morphing attacks \cite{fu_biosig_morph_2021}. Such observations suggest that the eye regions comprise more subtle and discriminative information for various face-related tasks.
% add more motivation and plot some heatmap figures 
As a result, we propose to weigh the predicted feature map regionally instead of just calculating the overall mean score. In our experiment, the weight of the eye region (including both the eyes and eyebrow region) is set to 0.6, while the weight for the face mask region is 0.1 and 0.3 for other regions, motivated by the observations in \cite{masked_face_pad}. Figure \ref{fig:method} illustrates the RW in the test phase, and the lighter color in the regional weight map refers to higher weight. The output feature map is multiplied with the regional weight map by the Hadamard product. Finally, a mean value of the weighted feature map is computed as the final decision score.

%%%%%%%%%%%%%%%%%%%%%%%%%%%%%%%%%%%%%%%%%%%%%%%%%%%%%%%%%%%%%%%%%%%%%%%%%%%%%%%%

\section{Experimental results} % ready for Naser
\label{sec:results}
In this section, we first introduce the used CRMA database, the detailed experimental setups, and the evaluation metrics. Then, the experimental results are described from two aspects, an extensive step-wise ablation study and the comparison with established PAD solutions. Finally, a discussion about possible future work is presented.

\subsection{Database}
\label{ssec:database}
In this work, we target the problem of the largely affected PAD performance when dealing with masked faces. The used dataset is the CRMA database \cite{masked_face_pad} which contains: 1) both unmasked (BM0) and masked (BM1) bona fide samples collected in a realistic scenario \cite{DBLP:conf/biosig/DamerGCBKK20,IET_B_Mask}, 2) conventional replay and print PAs created from faces not wearing a mask (AM0), 3) replay and printed PAs created from masked face images (AM1), and 4) partial attack where the unmasked printed/replayed faces are covered with real masks (AM2) (as shown in Fig.~\ref{fig:image_examples}). The bona fide data collected by Damer \etal \cite{DBLP:conf/biosig/DamerGCBKK20, IET_B_Mask} were adopted for investigation of the effect of wearing a mask on face verification performance, while the attack samples created by Fang \etal \cite{masked_face_pad} were used for the exploration of the effect of face mask on face PAD performance. The CRMA database comprises 423 bona fide videos and 12690 attack videos. Their results \cite{masked_face_pad} pointed out that the CRMA database is a challenging PAD database due to different face masks, multiple capture sensors, and various capture distances. In our experiment, we strictly follow the second protocol as defined in \cite{masked_face_pad}. The training, development, and test sets comprise unmasked and masked bona fide and all types of attack samples (BM0, BM1, AM0, AM1, AM2) (the number of samples in each set is detailed in \cite{masked_face_pad}). The identities in these three sets are disjoint to avoid identity-related biases.

\subsection{Experimental setup}
We following the implementation details described in \cite{deeppix_19,masked_face_pad,mixfacenet}. First, a face was detected and cropped by MTCNN \cite{DBLP:journals/spl/ZhangZLQ16}. Then, the 68 facial landmarks were detected by Dlib library \cite{dlib09}. Based on the jaw landmarks, a pixel-wise label was generated for the partial attack (AM2), in which the mask area is labeled as bona fide, and the rest of the face is labeled as an attack. For bona fide samples (BM0 and BM1), the values in this binary mask are all set to one (bona fide), while they are all set to zero for the other attack samples (AM0 and AM1). Finally, the input face image and the generated binary mask were used jointly while applying augmentation and resized to $224 \times 224 \times 3$ pixels and $14 \times 14$ pixels, respectively. 
In our experimental result discussion, we use PAL-RW$_{DeepPixBis}$ and PAL-RW$_{MixFaceNet}$ to indicate our PAL-RW solution by utilizing DeepPixBis and MixFaceNet network backbones, respectively.
For PAL-RW$_{DeepPixBis}$, the same augmentation techniques (horizontal flip and random jitter with the probability of 0.5) of DeepPixBis \cite{deeppix_19} were used in the training phase. The combined training loss was minimized by Adam Optimizer with the learning rate of $10^{-4}$ and the weight decay of $10^{-5}$. For PAL-RW$_{MixFaceNet}$, we employed the above augmentation techniques and the SGD optimizer with the learning rate of $10^{-2}$ and the weight decay of $5^{-3}$. The Exponential learning rate scheduler was adopted with the gamma of $0.995$. 
To further avoid overfitting, we applied class weight due to the unbalanced data and an early stopping technique with the maximum epoch of 100 and the stop patience of 15 for both models training processes. 
In the test phase, a regional weight map was generated based on the detected facial landmarks. The prediction score of each frame was a mean value of the regional weighted feature map. Finally, a final score for each video was computed by averaging the scores of the processed frames.
To put the achieved results in the perspective of the performance of well established PAD solutions, and not only as an ablation study, we adapt the selected solutions in \cite{masked_face_pad}. These solutions and implementation are detailed in \cite{masked_face_pad} and they include the following PAD solutions: LBP \cite{pad_lbp_2011}, CPqD \cite{pad_competition},  $\mathrm{Inception_{FT}}$ \cite{inception_v3}, $\mathrm{Inception_{FTS}}$ \cite{inception_v3}, $\mathrm{FASNet_{FT}}$ \cite{FASNet}, $\mathrm{Inception_{TFS}}$ \cite{FASNet}, DeepPixBis \cite{deeppix_19}. The results in Table \ref{tab:methods} of these methods are based on the reported results in \cite{masked_face_pad}.
All the experiments in this work, including the proposed method, the ablation study, and the list of above reported established PAD methods, use only the training data split of the second protocol of the CRMA dataset \cite{masked_face_pad} for their training, and the evaluation only used the testing data split of this protocol, and the thresholds are generated from the development data (the testing, training, and development are identity disjoint). 
%TRAINMING TESTING DATA....the second protocol as defined in \cite{masked_face_pad}. 

\subsection{Evaluation metrcis}
To evaluate the  performance of different PAD algorithms, we report the results based on the ISO/IEC 30107-3 \cite{ISO301073} standard metrics: \textit{Attack Presentation Classification Error Rate} (APCER) and \textit{Bona fide Presentation Classification Error Rate} (BPCER). The APCER is the proportion of attack images incorrectly classified as bona fide samples in a specific scenario, and the BPCER is the proportion of bona fide images misclassified as attacks in a specific scenario. APCER and BPCER values reported in the test set are based on a pre-computed threshold in the development set. To compare our results with other PAD methods in \cite{masked_face_pad}, we use a BPCER at 10\% on all data in the development set for obtaining the threshold (denoted as $\tau_{BPCER10}^{all}$) as defined in  \cite{masked_face_pad}. In addition, \textit{Average Classification Error Rate} (ACER) corresponding to the average value of the BPCER and APCER of the complete testing data is used for the evaluation of the overall performance. Moreover, to build a realistic ablation study where the behavior of the PAD on masked data is still unknown, we compute the threshold where the BPCER value at 10\% on only unmasked bona fide and attacks of the development set ($\tau_{BPCER10}^{unmask}$), and report the APCER, BPCER, and ACER on this threshold in the ablation study. Furthermore, Receiver Operating Characteristic (ROC) curves are demonstrated for further analysis on overall PAD performance. 

%for CR, add a comment about the weighting of the ROC.

\subsection{Experimental results}

\subsubsection{Ablation Study}
\label{sssec:ablation}

\begin{table*}[htbp!]
\begin{center}
\resizebox{0.99\textwidth}{!}{
\begin{tabular}{ccc|cc|ccc|ccc|c}
\hline
\multirow{2}{*}{Backbone} & \multirow{2}{*}{RW} & \multirow{2}{*}{PAL} & \multicolumn{2}{c|}{BPCER (\%)} & \multicolumn{3}{c|}{APCER (Print) (\%)} & \multicolumn{3}{c|}{APCER (Replay) (\%)} & \multirow{2}{*}{ACER (\%)} \\ %\cline{2-9} 
& & & BM0 & BM1 & AM0 & AM1 & AM2 & AM0 & AM1 & AM2 &  \\ \hline \hline
DeepPixBis & & & 63.16 & 64.04 & 0.00 & 0.00 & 0.00 & 0.00 & 0.64 & 0.00 & 29.47 \\
DeepPixBis & $\surd$ & & 35.09 & 41.23 & 0.00 & 0.10 & 1.17 & 0.19 & 1.95 & 0.58 & 18.58 \\
DeepPixBis & & $\surd$ & 42.11 & 51.75 & 0.00 & 0.00 & 0.00 & 0.00 & 0.44 & 0.00 & 23.35 \\
DeepPixBis & $\surd$ & $\surd$ & 26.32 & 29.82 & 0.00 & 0.19 & 1.17 & 0.00 & 1.32 & 0.29 & 14.81 \\ \hline
MixFaceNet & & & 5.26 & 7.89 & 12.09 & 9.75 & 11.70 & 21.83 & 17.68 & 14.33 & 13.80 \\
MixFaceNet & $\surd$ & & 5.26 &  7.89 & 11.50 & 9.36 & 11.70 & 21.64 &  17.15 & 14.33 & 13.75 \\
MixFaceNet & & $\surd$ & 12.28 & 7.89 & 4.29 & 4.39 & 8.19 & 17.54 &12.36 & 9.06 & 13.10 \\
MixFaceNet & $\surd$ & $\surd$ & 8.77 & 8.77 & 4.09 & 6.24 & 7.02 & 22.03 & 17.34 & 17.54 & 12.00 \\ \hline
\end{tabular}}
\end{center}
\caption{The PAD performance of the different step-wise ablation experiments using DeepPixBis and MixFaceNet as backbone network architectures on the CRMA database. Here, the used PAD decision threshold is the one scoring a BPCER 10\% on only unmasked data in the development set. The results show the individual and joint benefits of our PAL and RW components on PAD performance.}
\label{tab:ablation_results}
\vspace{-5mm}
\end{table*}

\begin{figure}[htb]
\begin{center}
\includegraphics[width=0.95\linewidth]{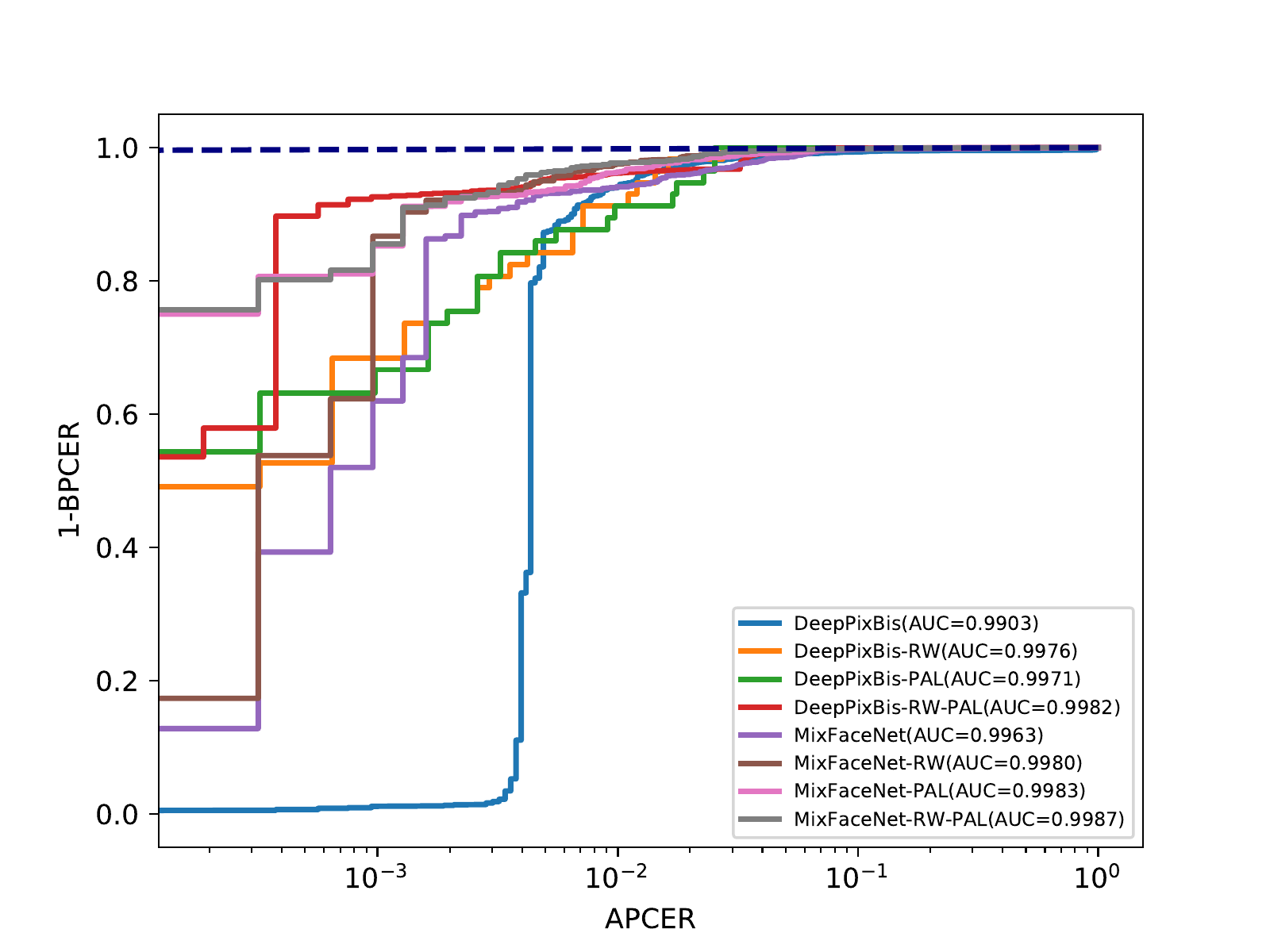}
\end{center}
\caption{ROCs of different solution steps of the ablation experiments in the CRMA database. Adding any one of the proposed modules (PAL or RW) improves the overall PAD performance. Moreover, experiments using the MixFaceNet backbone exhibit higher AUC values than using DeepPixBis architecture.}
\label{fig:ablation_roc}
\vspace{-5mm}
\end{figure}

To further validate the usefulness of each component of our solution, we conduct the experiments by gradually adding PAL and RW modules. The results are shown in Table \ref{tab:ablation_results}. We report the APCER and BPCER values by using the threshold  $\tau_{BPCER10}^{unmasked}$ that is pre-computed on only unmasked data in the development to build a realistic ablation study where the behavior of the PAD on masked data is still unknown. As shown in Table \ref{tab:ablation_results}, adding any one of the two components (PAL or RW) to the backbones does improve the PAD performance (considering at the overall performance index metric ACER). Moreover, the contribution made by the RW component is more significant than PAL while using DeepPixBis \cite{deeppix_19} as a backbone. For example, the ACER value is reduced from 29.47\% obtained by the DeepPixBis backbone to 18.58\% achieved by adding the RW component. Note that both setups use the same trained model, only the output prediction maps are weighted differently. This finding confirms our assumption that stresses the importance of eye regions for PAD decisions.
When using MixFaceNet \cite{mixfacenet} as the backbone, the reduction in the classification error rates is slightly smaller than DeepPixBis. Nevertheless, the basic MixFaceNet achieves lower overall PAD performance (13.80\% ACER value) in comparison to the DeepPixBis with additional PAL and RW modules (14.81\% ACER value). Such results indicate that MixFaceNet architecture possesses not only lower computational complexity but also higher generalization ability. In addition, the ROC curves for the ablation experiments are shown in Figure \ref{fig:ablation_roc}. The red curve (DeepPixBis-RW-PAL) and the grey curve (DeepPixBis-RW-PAL) are on top of other curves. Overall, the PAL and RW components can both improve the PAD performance in the CRMA database.

\subsubsection{Comparison with established PAD solutions}
\label{sssec:sotas}

\begin{table*}[htb!]
%\footnotesize
\centering
\def\arraystretch{1.0}
\resizebox{\textwidth}{!}{%
\begin{tabular}{c|cc|ccc|ccc|c}
\hline
\multirow{3}{*}{Method} & \multicolumn{9}{c}{Threshold @ BPCER 10\% on all data in dev set} \\ \cline{2-10}
& \multicolumn{2}{c|}{BPCER (\%)} & \multicolumn{3}{c|}{APCER (Print) (\%)} & \multicolumn{3}{c|}{APCER (Replay) (\%)} & \multirow{3}{*}{ACER (\%)} \\ 
& BM0 & BM1 & AM0 & AM1 & AM2 & AM0 & AM1 & AM2 &  \\ 
& 54 vids & 108 vids & 486 vids & 972 vids & 162 vids & 972 vids & 1944 vids & 324 vids & \\ \hline \hline 
LBP \cite{pad_lbp_2011} & 26.32 & 11.40 & 31.38 & 44.44 & 36.84 & 36.74 & 34.39 & 28.95 & 26.33 \\  
$\mathrm{Inception_{FT}}$ \cite{inception_v3} & 1.75 & 7.02 & 35.28 & 30.80 & 11.70 & 54.09 & 52.17 & 10.23 & 23.85 \\
CPqD \cite{pad_competition} & 3.51 & 7.89 & 27.49 & 30.41 & 16.37 & 46.20 & 44.50 & 10.23 & 21.75 \\
$\mathrm{FASNet_{FT}}$ \cite{FASNet} & 1.75 & 17.54 & 10.72 & 12.77 & 5.85 & 30.60 & 28.09 & 3.80 & 16.85  \\  % 9.40
$\mathrm{Inception_{TFS}}$ \cite{inception_v3} & 8.77 & 18.42 & 0.78 & 1.56 & 2.34 & 3.90 & 5.23 & 2.63 & 9.40 \\
$\mathrm{FASNet_{TFS}}$ \cite{FASNet} & 14.04 & 29.82 & 4.09 & 3.41 & 9.36 & 4.69 & 2.88 & 3.80 & 14.15 \\ 
DeepPixBis \cite{deeppix_19} & 12.28 & 31.58 & 1.75 &	0.29 & 4.09 & 0.88 & 1.37 & 0.29 & 13.13 \\  \hline
PAL-RW$_{DeepPixBis}$ (ours) & 18.07 & 28.95 & 0.00 & 0.19 & 1.75 & 0.00 & 1.37 & 0.29 & 12.99 \\ 
%MixFaceNet &  &  &  &  & &  &  &  & \\ 
PAL-RW$_{MixFaceNet}$ (ours) & 7.02 & 4.39 & 8.58 & 8.28 & 10.53 & 15.98 & 11.33 & 17.54 & \textbf{8.51} \\ \hline
\end{tabular}}
\caption{The PAD performance of our proposed PAL-RW methods using two network backbones on the CRMA database. The first two columns represent bona fide samples, and the left columns represent different attack types. The APCER and BPCER value is determined by a pre-computed threshold. This threshold is achieved at fixed BPCER 10\% on all (masked and unmasked) data in the development set for comparison with other methods in \cite{masked_face_pad}. The number of videos (denoting vids) of each category is noted in the header. The bold number indicates the lowest ACER value. Our PAL-RW$_{MixFaceNet}$ method outperforms other methods. }
\label{tab:methods}
\vspace{-5mm}
\end{table*}

\begin{figure}[htb]
\begin{center}
\includegraphics[width=0.95\linewidth]{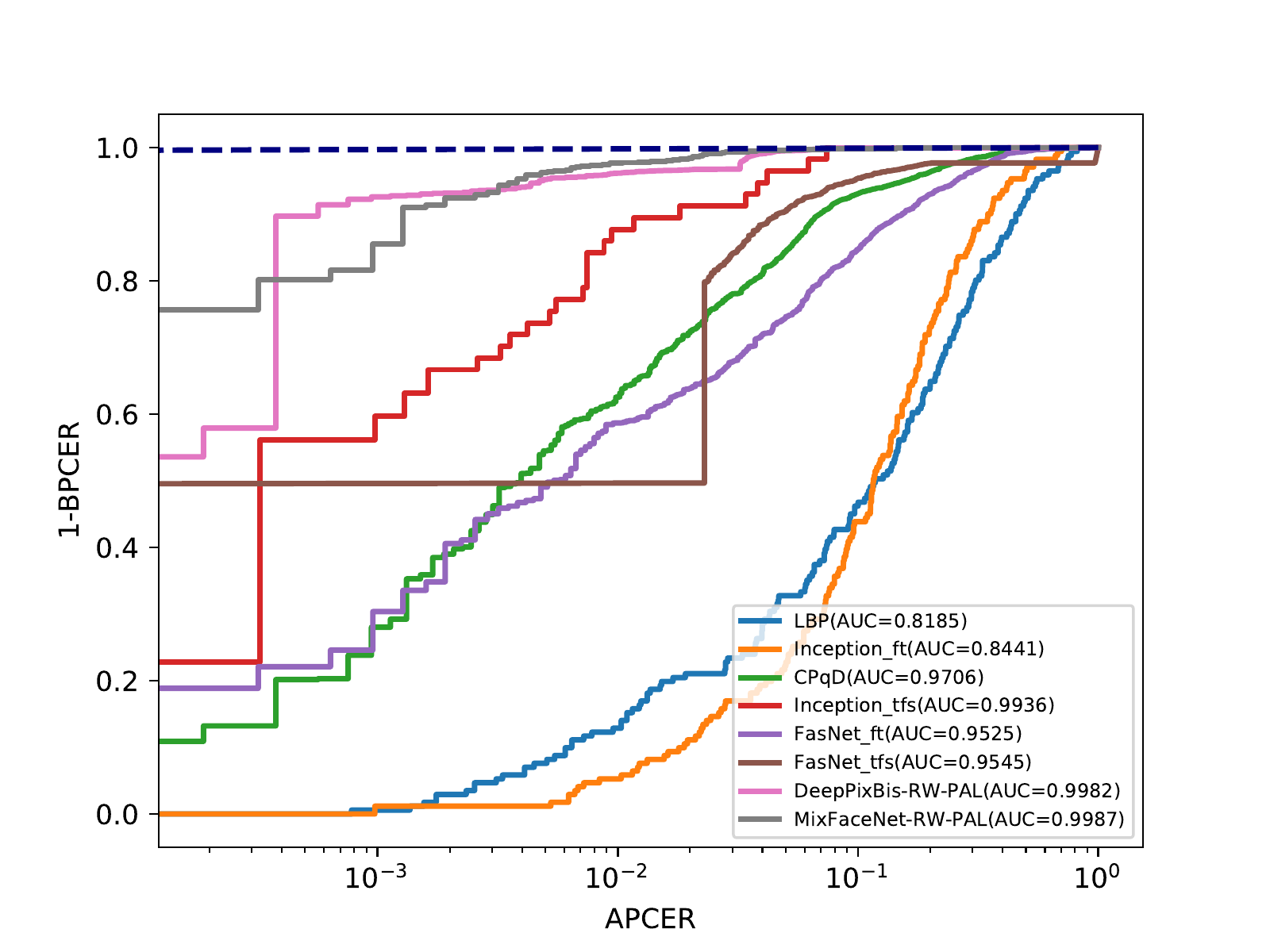}
\end{center}
\caption{ROCs of different PAD methods and our proposed PAL-RW methods on the CRMA database. PAL-RW$_{DeepPixBis}$ and PAL-RW$_{MixFaceNet}$ scores the lowest BPCER values (especially at low APCER) and achieve the largest areas under curves indicating the best overall performance. }
\label{fig:method_roc}
\vspace{-5mm}
\end{figure}

Table \ref{tab:methods} presents the results of the different investigated methods on the CRMA database and aims to put the performance achieved by our proposed PAL-RW solution in the perspective of the performance of established PAD solutions. The last two rows, PAL-RW$_{DeepPixBis}$ and PAL-RW$_{MixFaceNet}$, are results achieved by our PAL-RW methods. The other rows are the results of established PAD solutions reported in \cite{masked_face_pad}. The BPCER and APCER values in Table \ref{tab:methods} are determined by the threshold $\tau_{BPCER10}^{all}$, which follows the evaluation setup in \cite{masked_face_pad}. The bold number is the lowest ACER value indicating the best overall performance. As shown in Table \ref{tab:methods}, our PAL-WR method improves the overall PAD performance. For example, the ACER value decreases from the 13.13\% achieved by DeepPixBis to 12.99\% obtained by PAL-RW$_{DeepPixBis}$. In addition, the APCER value of the partial print attack (AM2) decreases from 4.09\% to 1.75\% when comparing the results of the DeepPixBis and the PAL-RW$_{DeepPixBis}$. This finding indicates that fine-grained partial attack labels are helpful for the improvement of  PAD performance under such circumstances. 
Moreover, PAL-RW$_{MixFaceNet}$ achieves the best overall performance (8.51\% ACER value). Note that the only difference between PAL-RW$_{MixFaceNet}$ and PAL-RW$_{DeepPixBis}$ methods is the backbone network architecture. Hence, the lower ACER value obtained by PAL-RW$_{MixFaceNet}$ indicates the efficiency of MixFaceNet \cite{mixfacenet} and rationalizes our choice of this efficient backbone based on the different sizes of convolutional kernels (thus the different capture levels of attack clues). 
In addition to comparing the results at a specific operation point, we also present the ROC curves for further observation on a wide range of decision thresholds. Figure \ref{fig:method_roc} illustrates the performance of different PAD methods. The pink (PAL-RW$_{DeepPixBis}$) and grey curves (PAL-RW$_{MixFaceNet}$) possess significantly larger areas under the curves than other methods and score lower BPCER values than the baseline methods, especially at low APCER values, which is consistent with the observation of Table \ref{tab:methods}.

\subsection{Discussions}
\label{ssec:discussion}
In this work, we demonstrate that the proposed PAL and RW components do substantially, and on multiple backbone networks, enhance the accuracy of PAD decisions when facing masked faces.
We also show that the proposed PAL-RW exhibits better generalization when dealing with masked faces than other established PAD methods. One of the advantages of our proposed method is that PAL-RW is not related to the network structure or training strategy and thus can be easily incorporated into any custom-designed network. Even though the proposed method is well-suited for masked face attacks, the PAL-RW method still has several limitations and can be improved in the future. First, the ground truth for partially masked attacks (AM2) is roughly generated based on 68 face landmarks and is only suitable for attacks with face masks. Therefore, it is worthwhile to produce accurate partial attack annotations (including other types of partial attacks) either manually or specifically designed. The fine-grained ground truth will enhance the generalization ability of PAD to unknown attacks. Second, in our case, the region weight map is set manually for all types of attacks, which is sub-optimal for the final PAD decision. One possible future work is to automatically perform regional weighted inference for different attack types, such as utilizing the position attention map. 
A partially masked face attack can be considered as an occlusion PAD problem.  Although current PAD algorithms have achieved good performance on 2D attacks  (print/replay) or 3D mask attacks, the occlusion PAD problem is still understudied. For example, the relevant partial attack data are insufficient. We stress that building a partial attack database is necessary for improving the generalizability of models to unknown attacks.

%%%%%%%%%%%%%%%%%%%%%%%%%%%%%%%%%%%%%%%%%%%%%%%%%%%%%%%%%%%%%%%%%%%%%%%%%%%%%%%%
%Naser here

\section{CONCLUSIONS} % NASER DONE
\label{sec:conclusion}
Recent studies have shown that PAD behavior is strongly affected by masked faces, whether as attacks or bona fide. 
%The spoof faces with protective face masks reduce the PAD performance. 
Driven by that observation, in this paper, we present a solution to target the masked presentation attacks, especially partially covered attacks, by proposing both the partial attack supervision and the regional weighted inference. The PAL is motivated by the pixel-wise supervision where each pixel in the binary mask can be considered as the label of several image patches. Hence, for the partially masked attack, the real face mask region is segmented and annotated as bona fide, the unmasked face region is labeled as an attack. Our proposed RW module used in the inference phase is inspired by previous observations stating that the eye regions contribute the most to face-related tasks, such as face recognition,  face presentation attacks, or morphing attacks. As a result, the output feature map is weighted regionally by giving the upper face region a relatively higher weight in the final PAD decision. The goal of PAL is to guide the neural network to better convergence while training.
Meanwhile, RW is to further improve the generalization ability of the model in the inference phase. We evaluate our proposed solutions on two different backbone neural network architectures, DeepPixBis and MixFaceNet, to demonstrate the experiments on the CRMA database. Our detailed ablation study shows the consistent benefits of both the PAL and RW components, separately and joint in a single solution.  Our PAL-RW$_{DeepPixBis}$ and PAL-RW$_{MixFaceNet}$ outperformed other established PAD methods when dealing with the possibility of masked faces in PAD decisions. As the PAL-RW focuses on the training ground truth and post-processing of PAD predictions, it thus can be easily incorporated into any neural architecture.

{\small
\bibliographystyle{ieee}
\bibliography{egbib}
}

\end{document}